\def\year{2022}\relax
\documentclass[letterpaper]{article} 
\usepackage{aaai22}  
\usepackage{times}  
\usepackage{helvet}  
\usepackage{courier}  
\usepackage[hyphens]{url}  
\usepackage{graphicx} 
\usepackage{natbib}  
\usepackage{caption} 
\DeclareCaptionStyle{ruled}{labelfont=normalfont,labelsep=colon,strut=off} 
\usepackage{algorithm}
\usepackage{algorithmic}

\usepackage{newfloat}
\usepackage{listings}

\usepackage{epsfig}
\usepackage{graphicx}
\usepackage{amsmath}
\usepackage{amssymb}

\usepackage{float} 
\usepackage{subfigure} 
\usepackage{amsmath}
\usepackage{booktabs}
\usepackage{multicol}
\usepackage{multirow}

\usepackage{amssymb}

\floatstyle{ruled}
\newfloat{listing}{tb}{lst}{}
\floatname{listing}{Listing}
\title{Modality-Adaptive Mixup and Invariant Decomposition for RGB-Infrared \\ Person Re-Identification}

\author{
Zhipeng Huang\textsuperscript{\rm 1}\equalcontrib, 
Jiawei Liu\textsuperscript{\rm 1}\equalcontrib, 
Liang Li\textsuperscript{\rm 2}, 
Kecheng Zheng\textsuperscript{\rm 1}, 
Zheng-Jun Zha\textsuperscript{\rm 1}\thanks{Corresponding Author}}
\affiliations {
    \textsuperscript{\rm 1} University of Science and Technology of China \\
    \textsuperscript{\rm 2} Institute of Computing Technology, Chinese Academy of Sciences \\
    \{hzp1104,zkcys001\}@mail.ustc.edu.cn, \{jwliu6,zhazj\}@ustc.edu.cn, liang.li@ict.ac.cn
}

\usepackage{bibentry}

\begin{document}

\maketitle

\begin{abstract}
RGB-infrared person re-identification is an emerging cross-modality re-identification task, which is very challenging due to significant modality discrepancy between RGB and infrared images. In this work, we propose a novel modality-adaptive mixup and invariant decomposition (MID) approach for RGB-infrared person re-identification towards learning modality-invariant and discriminative representations. MID designs a modality-adaptive mixup scheme to generate suitable mixed modality images between RGB and infrared images for mitigating the inherent modality discrepancy at the pixel-level. It formulates modality mixup procedure as Markov decision process, where an actor-critic agent learns dynamical and local linear interpolation policy between different regions of cross-modality images under a deep reinforcement learning framework. Such policy guarantees modality-invariance in a more continuous latent space and avoids manifold intrusion by the corrupted mixed modality samples. Moreover, to further counter modality discrepancy and enforce invariant visual semantics at the feature-level, MID employs modality-adaptive convolution decomposition to disassemble a regular convolution layer into modality-specific basis layers and a modality-shared coefficient layer. Extensive experimental results on two challenging benchmarks demonstrate superior performance of MID over state-of-the-art methods.
\end{abstract}

\section{Introduction}
Person re-identification (Re-ID)~\cite{sun2018beyond,liu2016multi,liu2019adaptive} has attracted increasing attention recently, due to its widespread applications in automated tracking~\cite{wang2013intelligent,kim2017omnitrack} and activity analysis~\cite{aggarwal2011human,caba2015activitynet}, \textit{etc}. It aims at identifying a target pedestrian from a gallery set captured from multi-disjoint camera views. Person Re-ID is quite challenging due to large intra-class and small inter-class variations caused by background clutter, occlusion, dramatic variations in illumination, body pose, \textit{etc}. Most existing person Re-ID methods mainly focus on RGB images of pedestrians from visible cameras and formulate the task as a single-modality (RGB-RGB) matching problem. They have achieved remarkable progresses in recent years for addressing appearance discrepancy (large intra-class and small inter-class variations). However, visible (RGB) cameras can not provide useful appearance information under poor illumination environments (\textit{e.g.}, at night), which limits the applicability of person Re-ID in a real scenario.

To handle this issue, recent surveillance systems begin to be equipped with infrared (IR) cameras to facilitate night-time monitoring, which raises a new cross-modality matching task termed RGB-infrared person Re-ID~\cite{wu2017rgb}. RGB-infrared person Re-ID aims to find the corresponding IR (or RGB) images of the same person captured by other spectrum cameras, given a RGB (or IR) image of a target person. Compared with the conventional single-modality person Re-ID, it encounters prominent modality discrepancy derived from the different imaging processes between different spectrum cameras (RGB and infrared images are intrinsically heterogeneous, which have different wavelength ranges), apart from appearance discrepancy. The key solution for RGB-infrared person Re-ID is to bridge large modality gap, and learn modality-invariant and discriminative features from RGB and IR images.

Existing RGB-infrared person Re-ID approaches mainly concentrate on mitigating the inherent modality discrepancy at the pixel-level or the feature-level to extract cross-modality shared features. For alleviating modality discrepancy at the pixel-level, these methods commonly design complex generative adversarial models~\cite{dai2018cross,wang2019learning,wang2020cross,wang2019rgb} to perform image-to-image translation and generate fake counterpart images, which are difficult to optimize and inevitably introduce noisy generated samples due to the ill-posed infrared-to-RGB transforming. On the other hand, for mitigating modality discrepancy at the feature-level, these methods employ one-stream~\cite{wu2017rgb,li2020infrared} or two-stream networks~\cite{ye2018hierarchical,ye2018visible,luo2020dynamic,hao2019hsme,zhu2020hetero,ye2019bi} to extract modality-invariant features by several customized losses. Nevertheless, one-stream network based methods learn a common network model, which lacks the capacity of explicitly modeling individual modalities and neglects modality specific characteristics, leading to crucial information loss. The two-stream network based methods firstly utilize separate branch layers for each modality to abstract modality-specific information, and then use non-branched shared layers for projecting the modality-specific features into a common feature space. They totally separate the process of modeling modality-specific and modality-share information, and could damage the vital cross-modality shared semantic during extracting modality-specific features. Furthermore, all aforementioned methods attempt to directly handle such large modality discrepancy and align two modalities, which are sensitive to the parameters and difficult to converge.

\begin{figure*}[!t]
	\centering
	\includegraphics[width=1.0\textwidth]{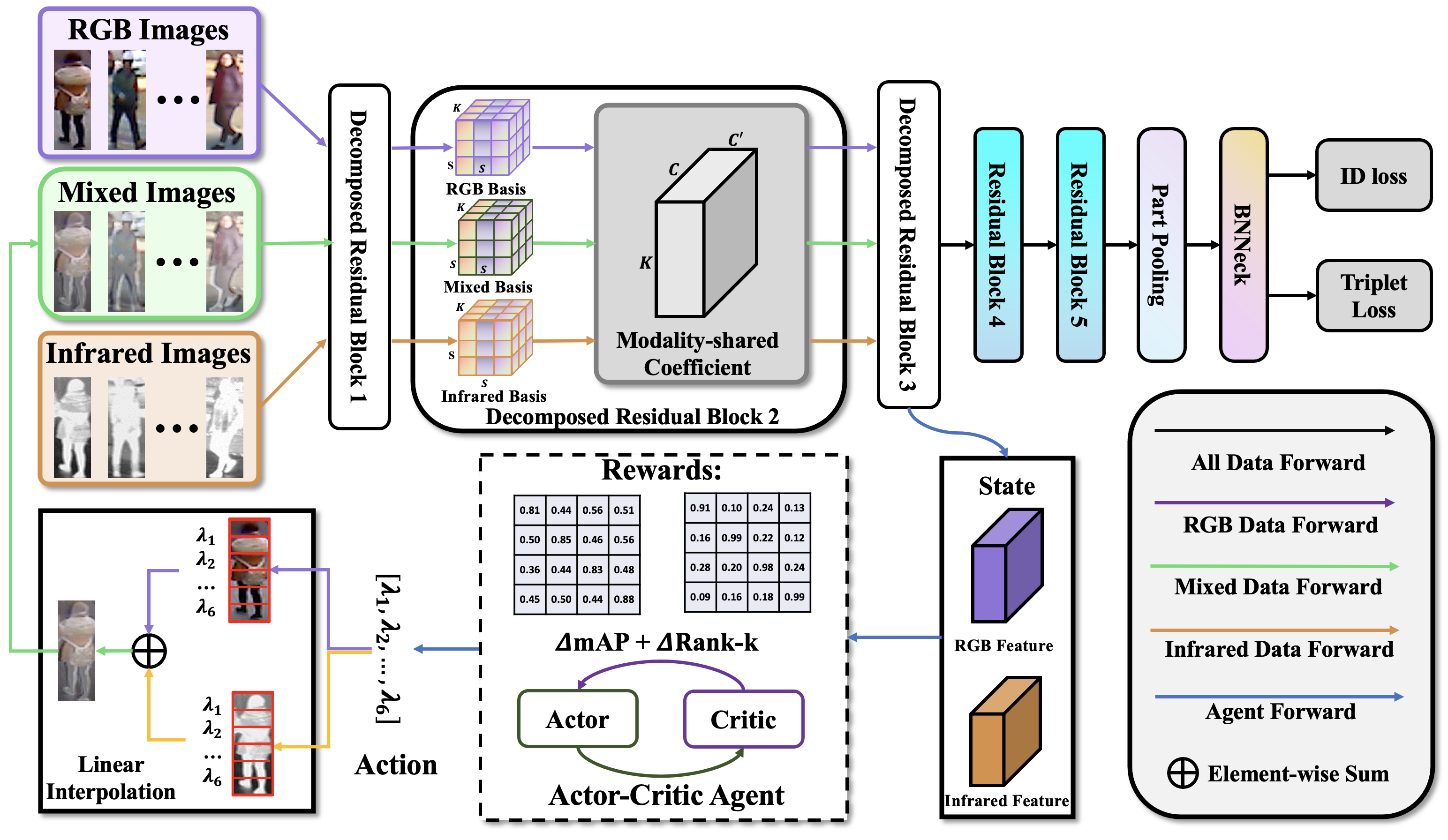}
	\caption{The overall architecture of the proposed MID. It consists of a modality-adaptive mixup module for generating appropriate mixed modality images and reducing the modality discrepancy at the pixel-level, and a decomposed convolution network for further shrinking the modality discrepancy at the feature-level to learn modality-invariant and discriminative representations.}
	\label{fig2}
\end{figure*}

In this work, we propose a novel modality-adaptive mixup and invariant decomposition (MID) approach for RGB-infrared person Re-ID towards learning modality-invariant and discriminative representations. MID designs a modality-adaptive mixup scheme (MAM) to generate appropriate mixed modality images and to reconcile modality gap at the pixel-level, according to the dynamical appearance and modality discrepancies between different RGB and IR images. MID then employs modality-adaptive convolution decomposition (MACD) to simultaneously counter modality discrepancy and enforce cross-modality shared semantics at the feature-level, towards facilitating cross-modality feature learning. Specifically, as shown in Figure \ref{fig2}, MAM is implemented by an actor-critic agent under a deep reinforcement learning framework. The state of the agent is the intermediate feature maps of RGB-infrared image pair, while the relevant action is the mixup ratio at a training step. With the joint optimization of the actor and critic networks, the suitability of the mixup ratio could be progressively boosted, driven by the supervision signal of reward that measures the relative performance improvement of person Re-ID by utilizing the generated mixed modality samples in the evaluation metrics. The actor-critic agent finally performs a dynamical and local linear interpolation policy between different regions of two modality images in a data-adaptive way, and generates augmented mixed modality samples with identity consistency. The mixed modality images mitigate modality discrepancy at the pixel-level and lead to a more continuous modality-invariant latent space. Moreover, MACD is designed to decompose a regular convolution layer into modality-specific basis layers and a modality-shared coefficient layer for handling the discrepancies among RGB, IR and mixed modalities. The former is in charge of modality intrinsic characteristics, while the latter enforces cross-modality shared decomposition coefficient to capture invariant semantics. ResNet-50 model~\cite{he2016deep} is equipped with MACD to learn aligned and discriminative features of pedestrians. Extensive experimental results on two datasets have shown the effectiveness of the proposed approach.  

Note that the related work XIV~\cite{li2020infrared} also generates X modality as an auxiliary modality to optimize the model. Nevertheless, XIV produces the limited X modality only from RGB images without considering the characteristics of IR images, failing to generate high-quality mixed modality samples for reducing modality gap and guaranteeing modality-invariant in a more continuous latent space.



The main contributions are summarized as follows: (1) We propose a novel modality-adaptive mixup and invariant decomposition for RGB-infrared person re-identification towards learning modality-invariant and discriminative representation. (2) We propose a modality-adaptive mixup scheme to generate suitable mixed modality samples for reconciling RGB and IR modalities at the pixel-level. (3) We design a modality-adaptive convolution decomposition to capture invariant visual semantics and shrink the modality discrepancy at the feature-level.

\section{Related work}

\subsection{RGB-Infrared Person Re-ID}
Existing RGB-infrared person Re-ID approaches mitigate the modality discrepancy at the pixel-level~\cite{dai2018cross,wang2020cross,fan2020modality} or the feature-level~\cite{wu2017rgb,li2020infrared,ye2018hierarchical,ye2018visible,luo2020dynamic,hao2019hsme}. For alleviating modality discrepancy at the pixel-level, Dai \textit{et al.}~\cite{dai2018cross} proposed a cross-modality generative adversarial network to handle the large-scale cross-modality metric learning problem. Wang \textit{et al.}~\cite{wang2020cross} proposed to generate cross-modality paired-images and perform both global set-level and fine-grained instance-level alignments. Fan \textit{et al.}~\cite{fan2020modality} proposed a modality-transfer generative adversarial network to generate a cross-modality counterpart of a source image in the target modality, for obtaining paired images and producing a general and robust unified feature embedding. Zhang~\textit{et al.}~\cite{zhang2021rgb} proposed a teacher-student GAN model (TS-GAN) to generate the corresponding fake IR images and guide the backbone to learn better feature. For mitigating modality discrepancy at the feature-level, Ye \textit{et al.}~\cite{ye2018hierarchical} proposed a hierarchical cross-modality matching model by jointly optimizing the modality-specific and modality-shared metrics. Ye \textit{et al.}~\cite{ye2018visible} proposed a dual-path network with a bi-directional dual-constrained top-ranking loss to learn discriminative feature representations. Luo~\textit{et al.}~\cite{luo2020dynamic} proposed a dynamic dual-attentive aggregation (DDAG) learning method by mining both intra-modality part-level and cross-modality graph-level contextual cues for RGB-infrared person Re-ID. Zhu~\textit{et al.}~\cite{zhu2020hetero} proposed a hetero-center loss to reduce the intra-class cross-modality variations and utilized a two-stream part-pooling-based model to obtain modality-invariant features.

\subsection{Domain Mixup}
\label{related:mixup}
Mixup is a recent data augmentation scheme to regularize deep learning models via \textit{global and random linear interpolations} between pairs of samples and their labels, which plays an important role in domain adaption \cite{zhang2017mixup, verma2019manifold,xu2020adversarial,zhong2020openmix}. For example, Zhang \textit{et al.} \cite{zhang2017mixup} proposed a simple data augmentation method termed MixUp, which randomly generates virtual training images by linearly interpolating two images and their corresponding labels. Panfilov \textit{et al.}~\cite{panfilov2019improving} investigated two regularization including mixup and adversarial unsupervised domain adaptation to improve the generalization of deep learning based knee cartilage segmentation model. Wu \textit{et al.}~\cite{wu2020dual} proposed a dual mixup regularized learning method for unsupervised domain adaptation, which jointly conducts category and domain mixup regularizations at the pixel-level to enhance the robustness of the model. Xu \textit{et al.}~\cite{xu2020adversarial} presented an adversarial domain adaptation with domain mixup (DM-ADA), which conduct on domain mixup on pixel and feature level to improve the robustness of models. Nevertheless, these existing mixup schemes are not suitable for RGB-Infrared Person Re-ID, which is elaborately analyzed in the next section.


\section{Method}

\subsection{Problem Definition and Overview}
Let $\boldsymbol{x}^{rgb}$ donates RGB image, and $\boldsymbol{x}^{ir}$ donate IR image. The RGB-IR training dataset is represented as $\boldsymbol{X} = \{\boldsymbol{x}^{rgb}_i, \boldsymbol{x}^{ir}_i, y_i\}_{i=1}^N$, where $N$ is the number of pedestrian images. Each image $\boldsymbol{x}^{rgb}$ or $\boldsymbol{x}^{ir}$ corresponds to an identity label $\boldsymbol{Y} =\{y_i\}_{i=1}^N$, where $y \in \{1,2,...,\mathcal{P} \}$ , and $\mathcal{P}$ refers to the number of persons. The objective is to learn a modality-invariant and discriminative representation for identifying the specific pedestrian from RGB and IR images. We propose a novel modality-adaptive mixup and invariant decomposition (MID) approach for RGB-infrared person Re-ID. As shown in Figure \ref{fig2}, MID firstly employs the modality-adaptive mixup scheme to synthesize augmented mixed modality images between RGB and IR images. The three modality images are then fed into a decomposed convolution network to learn modality-invariant and discriminative features, which is termed as $\mathcal{F}$. The first three residual blocks in $\mathcal{F}$ are denoted as $\mathcal{F}_1$, which are equipped with the designed MACD for absorbing modality variations and aligning invariant semantics across modalities. The decomposed convolution network is optimized with an identification loss and a center triplet loss for re-identification. 




\subsection{Modality-Adaptive Mixup Scheme}
\label{sec.mixup} 
Mixup is an effective data augmentation algorithm that regularizes deep learning models by linear interpolations between pairs of samples and their labels. Nevertheless, existing mixup scheme can not be directly applied to RGB-Infrared Person Re-ID for effectively reconciling RGB and IR modalities at the pixel-level. They commonly generate virtual samples via a simple \textit{global and random} linear interpolation policy. The mixup ratio for interpolating two samples is a scalar, and randomly sampled from a Beta$(\alpha, \alpha)$ distribution with a hyper-parameter $\alpha$. As the inter-modality discrepancy between different RGB and IR images dramatically changes, a random mixup ratio could produce low-quality mixed modality images, which is non-adjacent to real RGB and IR images of the same identity, resulting in manifold intrusion issue. Moreover, as RGB-Infrared Re-ID datasets lack paired RGB and IR images of the same pedestrian and have large intra-modality variations, a scalar mix ratio for globally interpolating RGB and IR images could corrupt the visual semantic and identity information of the mixed modality image, leading to performance degradation. 

For generating appropriate mixed modality images and guaranteeing the identity invariance, we propose a modality-adaptive mixup scheme to bridge the modality discrepancy at the pixel-level. It learns a dynamical and local linear interpolation between the different regions of cross-modality images in data-dependent fashion, which is formulated as Markov decision process and implemented by an actor-critic agent under a deep reinforcement learning (RL) framework~\cite{lillicrap2015continuous,li2020deep}. Specifically, the modality-adaptive mixup is computed as follows:
\begin{equation}
	\begin{split}
		\boldsymbol{x}^{mix}_{i,g} &= m_{i,g} \boldsymbol{x}^{rgb}_{i,g} + (1 - m_{i,g}) \boldsymbol{x}^{ir}_{i,g}, \\
		y^{mix}_{i} &= y^{rgb}_{i} = y^{ir}_{i},
	\end{split}
	\label{eq:mixup}
\end{equation}
where $m_{i,g}\in[0,1]$ is the mixup ratio learned from the actor-critic agent with RGB image $\boldsymbol{x}_i^{rgb}$ and IR image $\boldsymbol{x}_i^{ir}$. $\boldsymbol{x}_i^{rgb}$ and $\boldsymbol{x}_i^{ir}$ are divided into $G$ local regions of $\{\boldsymbol{x}_{i,g}^{rgb}\}_{g=1}^G$ and $\{\boldsymbol{x}_{i,g}^{ir}\}_{g=1}^G$ along horizontal axis, respectively. $\boldsymbol{x}^{mix}_{i,g}$ is the $g$-th local region of the generated mixed modality image. $y_i^{mix}$, $y_i^{rgb}$, $y_i^{ir}$ share the same identity.   

The mixup ratio dynamically adjusts based on the modality and appearance discrepancies between the corresponding local regions of RGB and IR images, which is performed by the actor-critic agent. The actor network $\mathcal{A}$ in the agent is employed to estimate the mixup ratio $\boldsymbol{m}_i$, and the critic network $\mathcal{Q}$ in the agent predicts the state-action value (Q-value). $\mathcal{A}$ is formulated as follows: 
\begin{equation}
	\boldsymbol{m}_i=\sigma\left(\boldsymbol{W}_{1} \delta(\boldsymbol{W}_{0}(\operatorname{Pool}( \operatorname{Conv}([\boldsymbol{F}_i^{rgb}, \boldsymbol{F}_i^{ir}]))))\right),
\end{equation}
where $\boldsymbol{F}_i^{rgb}$ denotes the intermediate feature map of $\boldsymbol{x}^{rgb}_{i}$ extracted from $\mathcal{F}_1$. $\sigma$ and $\delta$ denote the sigmoid function and rectified linear unit (ReLU) activation function, respectively. $\boldsymbol{W}_0$ and $\boldsymbol{W}_1$ denote two Fully Connected (FC) layers. $\operatorname{Conv}$ denotes a $3\times 3$ convolution layer, followed with a batch normalization layer and a ReLU layer. $\operatorname{Pool}$ indicates a global average pooling layer. The critic network in the actor-critic agent has a similar network architecture to the actor network. For guiding the critic network to predict the reliability of the mixup ratio, a reward $\mathcal{R}$ is designed as the supervision signal for optimization. The action, state and reward for the agent are defined as follows:  

\textbf{State.} The concatenation of the intermediate feature maps $[\boldsymbol{F}_i^{rgb}; \boldsymbol{F}_i^{ir}]$ of the RGB and IR images extracted from $\mathcal{F}_1$ is viewed as the state of the agent.

\textbf{Action.} The action of the agent is the mix ratio $\boldsymbol{m}_i \in \mathbb{R}^G$ used for linear interpolation bewteen RGB and IR images. It is a continuous vector.    

\textbf{Reward.} For a mini-batch input data $\{\boldsymbol{x}^{rgb}_{i}, \boldsymbol{x}^{ir}_{i}\}_{i=1}^{b}$ with $y^{rgb}_{i} = y^{ir}_{i}$, we can obtain the mixed modality images $\{\boldsymbol{x}_i^{mix}\}_{i=1}^{b}$ via dynamic and local linear interpolation in Eq.~(1). We then calculate the similarity matrix $[\boldsymbol{\mathcal{S}}^{rgb,ir}]_{i,j}= \langle\boldsymbol{f}^{rgb}_i, \boldsymbol{f}^{ir}_j\rangle$ between all RGB and IR images, and $[\boldsymbol{\mathcal{S}}^{mix,ir}]_{i,j} = \langle\boldsymbol{f}^{mix}_i, \boldsymbol{f}^{ir}_j\rangle$ between all mixed modality and IR images, where $\boldsymbol{f}_i$ denotes the learned pedestrian representation from the decomposed convolution network $\mathcal{F}$. $\langle\cdot,\cdot\rangle$ denotes Cosine Distance. Thus, the reward is defined as the relative performance improvement of re-identification by using the mixed modality images to enhance the similarity matrix, which is formulated as follows:   
\begin{equation}
	\begin{aligned}
		\mathcal{R}=&\mathcal{E}(\boldsymbol{\mathcal{S}}^{rgb,ir}+\boldsymbol{\mathcal{S}}^{mix,ir})-\mathcal{E}(\boldsymbol{\mathcal{S}}^{rgb,ir})\\
		&+\mathcal{E}(\boldsymbol{\mathcal{S}}^{ir,rgb}+\boldsymbol{\mathcal{S}}^{mix,rgb})-\mathcal{E}(\boldsymbol{\mathcal{S}}^{ir,rgb}), \\
	\end{aligned}
\end{equation}

\begin{equation}
	\begin{aligned}
		\mathcal{E}(\boldsymbol{\mathcal{S}})=\operatorname{mAP}(\boldsymbol{\mathcal{S}})+\sum_{k=1}^{K}\frac{1}{k}\operatorname{rank-k}(\boldsymbol{\mathcal{S}}),
	\end{aligned}
\end{equation}
where $\operatorname{mAP}$ and $\operatorname{rank-k}$ are the common evaluation metrics for RGB-Infrared Person Re-ID based on the similarity matrix \cite{ye2018hierarchical,ye2018visible}. $K$ denotes the number of the pedestrian images for one identity in each modality. We adopt the losses $\mathcal{L}_{\mathcal{A}}$, $\mathcal{L}_{\mathcal{Q}}$ to optimize the actor and the critic networks, respectively, which will be introduced in the following subsection. The data-dependent modality-adaptive mixup scheme explicitly exploits the data distribution of two modalities, and can learn appropriate interpolation policy to generate high-quality mixed modality images for reducing modality discrepancy at the pixel-level and facilitating a more continuous modality-invariant latent space.

\subsection{Modality-Adaptive Conv Decomposition}

\label{sec.decompose}



To bridge modality gap at the feature-level, previous works employs one-stream network or two-stream network architecture. Nevertheless, one-stream network based methods lack the capacity of explicitly modeling modality-specific characteristics, resulting in unnecessary information loss. Two-stream network based methods separate the process of modeling modality-specific and modality-share information, vitiating the significant cross-modality shared semantic during extracting modality-specific features. They also require more training parameters. 

To address these issues, we propose modality-adaptive convolution decomposition and design a decomposed convolution network $\mathcal{F}$. The decomposed convolution network is built on ResNet-50 model. The modality-adaptive convolution decomposition approximates convolution filters as a linear combination of a small set of dictionary bases, for simultaneously countering modality discrepancy and enforcing cross-modality shared semantic at the feature-level. It is applied to the convolution layers of first three residual blocks $\mathcal{F}_1$ in the decomposed convolution network, each of which is decomposed into modality-specific basis layers and a modality-shared coefficient layer, while both of them remain convolutional. Specifically, $\boldsymbol{W}$ denotes the convolution filters $S \times S \times C_{in} \times C_{out}$ of a convolution layer, where $S$ is the spatial size of the filters, $C_{in}$ and $C_{out}$ denotes the number of input channel and output channel. We decompose $\boldsymbol{W}$ into the modality-specific dictionary bases $\boldsymbol{\alpha} \in \mathbb{R}^{S \times S \times K}$ and the common coefficient across modalities $\boldsymbol{\Psi} \in \mathbb{R}^{K \times C_{in} \times C_{out}}$. In detail, each decomposed convolution layer has three independent modality-specific dictionary bases $\boldsymbol{\alpha}^{rgb},\boldsymbol{\alpha}^{ir},\boldsymbol{\alpha}^{mix}$, and the common coefficient $\boldsymbol{\Psi}$ across modalities:
\begin{equation}
	\boldsymbol{W}^* = [\boldsymbol{\alpha}^{\{rgb\}}\boldsymbol{\Psi}, \boldsymbol{\alpha}^{\{ir\}}\boldsymbol{\Psi}, \boldsymbol{\alpha}^{\{mix\}}\boldsymbol{\Psi}],
\end{equation}
The modality-specific dictionary bases are independently learned from the corresponding modality images to model the modality variations. They convolve spatially each individual input feature channel $[\boldsymbol{F}]_{1:C_{in}} \in \mathbb{R}^{H\times W}$ for modality discrepancy correction. The common coefficient is learned from all the three modality images, and performs $1 \times 1$ convolution for weighting sum the corrected output feature channels, thus promoting cross-modality shared semantic. The decomposed convolution network takes RGB, IR and mixed modality images from the modality-adaptive mixup module as input, and effectively handles the large modality gap at the feature-level for learning modality-invariant features. The network outputs global and local features $\boldsymbol{f}^{rgb}_i$,  $\boldsymbol{f}^{ir}_i$, $\boldsymbol{f}^{mix}_i$ of three modalities by part pooling operation \cite{sun2018beyond}.

\subsection{Loss Function and Optimization}
\label{sec.loss}

We adopt the identification loss $L_{id}$ with label smoothing regularization \cite{ainam2019sparse} and the center triplet loss $L_{ct}$ \cite{he2018triplet, hardtripletcenter} to optimize the decomposed convolution network for re-identification. The identification loss is calculated as follows:
\begin{equation}
	L_{id}=\sum_{i=1}^{\mathcal{P}}-q_i\log\left(p_i\right) 
	,\; q_i=\left\{\begin{array}{cl}1-\frac{\mathcal{P}-1}{\mathcal{P}} \xi, & y=i \\ \frac{\xi}{\mathcal{P}}, & y \neq i\end{array}\right.
\end{equation}
where $\mathcal{P}$ is the number of identities in the training set, $y$ is the ground-truth ID and $p_i$ denotes the ID prediction logits of the $i^{th}$ class. $\xi$ is the smoothing parameter, which is beneficial to prevent the network from over-fitting to training IDs. The three identification losses $L_{id}^{rgb}$ , $L_{id}^{ir}$ , $L_{id}^{mix}$ are used to supervise the global and local features of three modalities $\boldsymbol{f}^{rgb}_i$,  $\boldsymbol{f}^{ir}_i$, $\boldsymbol{f}^{mix}_i$, respectively. For the center triplet loss, we randomly sample $P$ identities and $K$ images of each identity for RBG and IR modality to form a mini-batch with $PK$ images termed $\{\boldsymbol{x}^{rgb}_{p,k}, \boldsymbol{x}^{ir}_{p,k}, y_{p}\}_{p=1, k=1}^{P,K}$. We calculate the feature centers of each pedestrian for the three modality images in a mini-batch, $\boldsymbol{c}^{\{rgb,ir,mix\}}_p = \frac{1}{K} \sum_{k=1}^K \boldsymbol{f}^{\{rgb,ir,mix\}}_k$. Thus, the center triplet loss is defined as follows:
\begin{equation}
	\begin{aligned}
		L_{ct}^{\alpha,\beta}=
		&\sum_{i=1}^{P}[\rho+
		\|c^{\alpha}_p-\boldsymbol{c}^{\beta}_p\|_{2}-\min\limits_{\begin{subarray}{\boldsymbol{c}}n\in\{\alpha,\beta\} \\ j\neq p\end{subarray}}\|\boldsymbol{c}^{\alpha}_p-\boldsymbol{c}^{n}_{j}\|_{2}]_{+} 
		& \\
		&+\sum_{i=1}^{P}[\rho+
		\|\boldsymbol{c}^{\beta}_p-\boldsymbol{c}^{\alpha}_p\|_{2}-\min\limits_{\begin{subarray}{\boldsymbol{c}}n\in\{\alpha,\beta\} \\ j\neq p\end{subarray}}\|\boldsymbol{c}^{\beta}_p-\boldsymbol{c}^{n}_{j}\|_{2}]_{+},
	\end{aligned}
	\label{loss:ct}
\end{equation}
where $\rho$ is a margin parameter, $[z]_+ = max(z, 0)$, $\alpha,\beta \in \{rgb,ir,mix\}$ which denote two different modalities. We utilize three center triplet losses $L_{ct}^{rgb,ir}, L_{ct}^{rgb,mix}, L_{ct}^{ir,mix}$ to supervise the three modality features $\boldsymbol{f}^{rgb}_i$,  $\boldsymbol{f}^{ir}_i$, $\boldsymbol{f}^{mix}_i$. Therefore, the total loss $L_{dcn}$ for the decomposed convolution network is the combination of these identification and center triplet losses:
\begin{equation}
	\begin{aligned}
		L_{dcn} = &\lambda_{1}L_{ct}^{rgb,ir} + \lambda_{2}L_{ct}^{rgb,mix} + \lambda_{3}L_{ct}^{ir,mix} \\  &+  \lambda_{4}L_{id}^{rgb} + \lambda_{5}L_{id}^{ir}+ \lambda_{6}L_{id}^{mix},
		\label{eq.loss}
	\end{aligned}
\end{equation}
where $\lambda_{1-6}$ is the trade-off parameters. 

Different from standard reinforcement learning algorithm, the proposed actor-critic agent does not have explicit sequential relationship along different training steps. The action of the interpolation policy between RGB and IR images is conditioned on the state of their intermediate features in a one-shot fashion, which is essentially a one-step Markov decision process. The action space is continuous, thus the optimal action could be found by gradient ascent method following the solution of continuous Q-value prediction \cite{lillicrap2015continuous,li2020deep}. The loss for actor network is defined as follows:
\begin{equation}
	L_{\mathcal{A}}=-\mathcal{Q}(\mathcal{A}([\boldsymbol{F}^{rgb}, \boldsymbol{F}^{ir}]), [\boldsymbol{F}^{rgb}, \boldsymbol{F}^{ir}]),
\end{equation}
The actor network $\mathcal{A}$ is updated to achieve higher Q-value, which implies higher rank-$k$ accuracy and mAP for person Re-ID. The critic network is optimized to predict an accurate Q-value estimation, thus MSE loss is employed as follows:
\begin{equation}
	L_{\mathcal{Q}}=\|\mathcal{Q}(\mathcal{A}([\boldsymbol{F}^{rgb}, \boldsymbol{F}^{ir}]), [\boldsymbol{F}^{rgb}, \boldsymbol{F}^{ir}])-\mathcal{R}\|^{2},
\end{equation}
For the entire MID, we combine the advantage of supervised and reinforcement learning, alternately optimizing the decomposed convolution network and the actor-critic agent. 


\section{Experiments}

\begin{table*}[!ht]
	\centering
	\caption{Performance (\%) comparison to the state-of-the-art methods on RegDB and SYSU-MM01 datasets.}
	\resizebox{\linewidth}{!}{
	\begin{tabular}{l|cc|c|cc|c|cc|c|cc|c}
		\hline
		\hline
		\multicolumn{1}{c|}{\multirow{3}[0]{*}{Methods }} & \multicolumn{6}{c|}{RegDB}                                     & \multicolumn{6}{c}{SYSU-MM01} \\ 
		& \multicolumn{3}{c}{Visible to Thermal} & \multicolumn{3}{c}{Thermal to Visible} & \multicolumn{3}{|c}{All}       & \multicolumn{3}{c}{Indoor} \\
		& \multicolumn{1}{c}{r1} & \multicolumn{1}{c}{r10} &  \multicolumn{1}{c}{mAP} & \multicolumn{1}{c}{r1} & \multicolumn{1}{c}{r10}  & \multicolumn{1}{c|}{mAP} & \multicolumn{1}{c}{r1} & \multicolumn{1}{c}{r10}  & \multicolumn{1}{c}{mAP} & \multicolumn{1}{c}{r1} & \multicolumn{1}{c}{r10}  & \multicolumn{1}{c}{mAP} \\
		\hline
		\hline
		Zero-Pad~\cite{wu2017rgb} & 17.75 & 34.21  & 18.90 & 16.63 & 34.68  & 17.82 & 14.80 & 54.12           & 15.95 & 20.58 & 68.38 & 26.92 \\
		HCML~\cite{ye2018hierarchical}     & 24.44 & 47.53  & 20.80 & 21.70 & 45.02  & 22.24 & 14.32 & 53.16           & 16.16 & 24.52 & 73.25  & 30.08 \\
		MAC~\cite{ye2019modality}      & 36.43 & 62.36  & 37.03 & 36.20 & 61.68  & 39.23 & 33.26 & 79.04           & 36.22 & 36.43 & 62.36 & 37.03 \\
		MSR~\cite{feng2019learning}      & 48.43 & 70.32  & 48.67 &  -    &   -    &   -   & 37.35 & 83.40           & 38.11 & 39.64 & 89.29  & 50.88 \\
		D-HSME~\cite{hao2019hsme}   & 50.85 & 73.36  & 47.00 & 50.15 & 72.40  & 46.16 & 20.68 & 62.74           & 23.12 &   -   &   -    & - \\
		EDFL~\cite{liu2020enhancing}     & 52.58 & 72.10  & 52.98 & 51.89 & 72.09  & 52.13 & 36.94 & 85.42           & 40.77 &   -   &  -      & - \\
		AlignGAN~\cite{wang2019rgb} & 57.90 &    -   & 53.60 & 56.30 &   -    & 53.40 & 42.40 & 85.00           & 40.70 & 45.90 & 87.60  & 54.30 \\
		TS-GAN~\cite{zhang2021rgb} & - & - & - & - & - & - & 49.8 & 87.3 & 47.4 & 50.4 & 90.8  & 63.1 \\
		XIV~\cite{li2020infrared}      & 62.21 & 83.13  & 60.18 &   -   & -      &  -    & 49.92 & 89.79           & 50.73 &   -   &  -      & - \\
		DDAG~\cite{luo2020dynamic},     & 69.34 & 86.19  & 63.46 & 68.06 & 85.15  & 61.80 & 54.75 & 90.39           & 53.02 & 61.02 & 94.06 & 67.98 \\
		Hi-CMD~\cite{choi2020hi}   & 70.93 & 86.39  & 66.04 &   -   & -      &  -    & 34.94 & 77.58           & 35.94 &   -   &   -     & - \\
		TSLFN~\cite{zhu2020hetero}   & - & -  & - & - & -  & - & 56.96 & 91.50           & 54.95 & 59.74 & 92.07  & 64.91 \\
		HAT~\cite{ye2020visible}      & 71.83 & 87.16  & 67.56 & 70.02 & 86.45  & 66.30 & 55.29 & \underline{92.14}          & 53.89 & 62.10 & 95.75  & 69.37 \\
		$G^2$DA~\cite{wan2021g} & 71.72 & 87.13 & 65.90 & 69.50 & 84.87 & 63.88 & \underline{57.07} & 90.99  & 55.05 & \underline{63.70} & 94.06 & 69.83 \\
		NFS~\cite{chen2021neural} & \underline{80.54} & \underline{91.96} & \underline{72.10} & \underline{77.95} & \underline{90.45} &  \underline{69.79}      & 56.91 & 91.34 & \underline{55.45} & 62.79 & \textbf{96.53} & \underline{69.79} \\
		\hline
		MID (ours)  & \textbf{87.45} & \textbf{95.73} & \textbf{84.85} & \textbf{84.29} & \textbf{93.44} & \textbf{81.41} & \textbf{60.27} & \textbf{92.90}  & \textbf{59.40} & \textbf{64.86} & \underline{96.12}  & \textbf{70.12} \\ \hline \hline
	\end{tabular}%
 	}
	\label{tab:state-of-the-art}%
	\vspace{-0.5cm}
\end{table*}%

\subsection{Experimental Setting}

\noindent \textbf{Datasets.} We evaluate the proposed MID using two public RGB-Infrared datasets: RegDB~\cite{nguyen2017person} and SYSU-MM01~\cite{wu2017rgb}. RegDB dataset contains 412 pedestrians. Each pedestrian has 10 visible images and 10 thermal images. Following the evaluation protocol~\cite{ye2018hierarchical,ye2018visible}, this dataset is randomly split into two parts, 206 identities for training and the other 206 identities for testing, with two different testing modes, \textit{i.e.}, visible to thermal mode and thermal to visible mode. The reported results are the average of 10 random training/test splits on RegDB dataset. SYSU-MM01~\cite{wu2017rgb} is the largest existing RGB-infrared dataset, which was captured with 4 visible and 2 infrared cameras. The training set contains 395 persons with 22,258 RGB images and 11,909 IR images, while the testing set contains 96 persons with 3,803 IR images and 301 RGB images. We adopt the all-search mode and indoor-search mode \cite{wu2017rgb, luo2019spectral} to evaluate the performance. 

\noindent\textbf{Evaluation Metrics.} The standard Cumulative Matching Characteristic (CMC) at Rank-$k$ and the mean Average Precision (mAP) are adopted as evaluation metrics.

\noindent\textbf{Implementation Details:} The proposed method is implemented by the PyTorch framework with one NVIDIA Tesla V100 GPU. Each mini-batch contains 96 images of 8 identities (each person has 4 RGB images, 4 IR images, and 4 generated mixed modality images). ResNet-50~\cite{he2016deep} model is adopted as the backbone network. Part-pooling~\cite{sun2018beyond} is added after the backbone. The first three residual blocks of ResNet-50 model are equipped with modality-adaptive convolution decomposition. The stride of the last convolution layer is set to 1. The margin $\rho$ is set to 0.3. The parameter $\mu$ and  $\xi$ are set to 1 and 0.1, respectively. The trade-off parameters $\lambda_{1,4,5}$ are set to 1, $\lambda_{2,3}$ are set to 0.5, and $\lambda_6$ is set to 0.1 in Eq.~(\ref{eq.loss}). We adopt Adam Optimizer to train the actor-critic agent. And we utilize the stochastic gradient descent (SGD) optimizer for MACD with the momentum of 0.9, the initial learning rate of 0.05, 0.02 on RegDB and SYSU-MM01 datasets, respectively. The learning rates decayed by 0.1 after 20 and 45 epochs. The whole MID framework is trained for 60 epochs on RegDB dataset which takes 1 hour, and for 100 epochs on SYSU-MM01 dataset which takes 6 hours.



\subsection{Comparison to State-of-the-Art Methods}


\textbf{RegDB:} We present a comparison the proposed MID with 11 state-of-the-art approaches on RegDB dataset in Table~\ref{tab:state-of-the-art}. Experiments conducted on RegDB dataset show that the proposed MID achieves the best performance under different testing modes over all state-of-the-art methods by large margins. For Visible to Thermal mode, MID achieves 87.45\% rank-1 accuracy and 84.85\% mAP, outperforming the 2nd best NFS~\cite{chen2021neural} by 6.91\% rank-1 accuracy and 12.75\% mAP, respectively. For Infrared to RGB mode, MID also obtains 84.29\% rank-1 accuracy and 84.29\% mAP, improving the 2nd best method NFS~\cite{chen2021neural} by 6.34\% rank-1 accuracy and 11.62\% mAP, respectively. The boosting demonstrates that the proposed MID is able to learn modality-invariant and discriminative features from cross-modality RGB and infrared images.  

\textbf{SYSU-MM01:} Table~\ref{tab:state-of-the-art} also reports the performance of the proposed MID with 11 state-of-the-art methods on SYSU-MM01 dataset. Experiments results show that MID obtains the best performance under both All-search and Indoor settings. For all-search mode,  MID achieves 60.27\% rank-1 accuracy and 59.40\% mAP, outperforming the 2nd best method NFS~\cite{wan2021g} by 3.20\% rank-1 accuracy and 4.35\% mAP score. For indoor-search mode,  MID also obtains the best rank-1 accuracy and mAP score. The comparisons demonstrate that MID can effectively reduce inter-modality and intra-modality discrepancies.



\subsection{Ablation Study}
\label{sec.ablation}

\begin{table}[!t]
	\footnotesize
	\begin{center}
		\caption{Ablation studies on the effectiveness of each component of the proposed MID.}
		\scalebox{1}{
			\centering
			\begin{tabular}{ccc|c|c|c|c}
				\hline
				\hline
				\multicolumn{1}{c}{\multirow{2}[0]{*}{$\mathcal{B}$}} & \multicolumn{1}{c}{\multirow{2}[0]{*}{$\mathcal{M}$}} & \multicolumn{1}{c|}{\multirow{2}[0]{*}{$\mathcal{D}$}} & \multicolumn{2}{c|}{RegDB} & \multicolumn{2}{c}{SYSU-MM01} \\ 
				&& & \multicolumn{1}{c}{r1} & \multicolumn{1}{c|}{mAP} & \multicolumn{1}{c}{r1} & \multicolumn{1}{c}{mAP}\\ \hline
				 $\checkmark$ & $\times$ &$\times$  & 76.34 & 70.81 & 49.46 & 49.32 \\
				 $\checkmark$ & $\checkmark$ &$\times$ & 83.43 & 79.13 & 56.22 & 57.12 \\ 
				 $\checkmark$ & $\checkmark$&$\checkmark$  & 87.45 & 84.85 & 60.27 & 59.40 \\
				\hline
				\hline
			\end{tabular}
		}
		\label{tab:module-analysis}
		\vspace{-0.5cm}
	\end{center}
\end{table}

\textbf{Effectiveness of each component of MID.} We conduct ablation studies on RegDB and SYSU-MM01 datasets to investigate the effectiveness of the components of MID in Table~\ref{tab:module-analysis}, including modality-adaptive mixup scheme ($\mathcal{M}$) and modality-adaptive convolution decomposition ($\mathcal{D}$). $\mathcal{B}$ denotes a vanilla ResNet-50 model trained with identity loss and normal triplet loss. When introducing the modality-adaptive mixup scheme, the performance is improved by 7.09\% rank-1 accuracy and 8.32\% mAP on RegDB dataset, and by 6.76\% rank-1 accuracy and 7.80\% mAP on SYSU-MM01 dataset. The boosting indicates the dynamical and local linear interpolation policy of $\mathcal{M}$ can generate appropriate mixed modality images and mitigate the inherent modality discrepancy at the pixel-level for promoting a more continuous modality-invariant latent space. By adding the modality-adaptive convolution decomposition, the performance is improved by 4.02\% rank-1 accuracy and 5.72\% mAP on RegDB dataset, and by 4.05\% rank-1 accuracy and 2.28\% mAP on SYSU-MM01 dataset. This demonstrates that the decomposed convolution network with $\mathcal{D}$ can capture invariant visual semantics and further shrink the modality discrepancy at the feature-level. 

\begin{table}[!t]
	\small
	\centering
	\caption{Influence of different mixup schemes for MID.}
	\begin{tabular}{c|cccc}
		\hline \hline
		\multicolumn{1}{c|}{\multirow{2}[0]{*}{Mixup Scheme}} & \multicolumn{2}{c}{RegDB} & \multicolumn{2}{c}{SYSU-MM01} \\ 
		& \multicolumn{1}{l}{r1} & \multicolumn{1}{l}{mAP} & \multicolumn{1}{l}{r1} & \multicolumn{1}{l}{mAP} \\
		\hline
		Fix & 83.96 & 77.83 & 50.43 & 48.64 \\
		Beta & 83.53 & 78.04 & 51.46 & 50.09 \\
		MAM & 87.45 & 84.85 & 60.27 & 59.40 \\
		\hline \hline
	\end{tabular}%
	\label{tab:mixup-policy}%
\end{table}%

\begin{table}[!t]
	\small
	\centering
	\caption{Influence of different number $G$ of the partitioned local regions for mixup.}
	
	\begin{tabular}{c|cccc}
		\hline \hline
		\multicolumn{1}{c|}{\multirow{2}[0]{*}{Mixup Scheme}} & \multicolumn{2}{c}{RegDB} & \multicolumn{2}{c}{SYSU-MM01} \\ 
		& \multicolumn{1}{l}{r1} & \multicolumn{1}{l}{mAP} & \multicolumn{1}{l}{r1} & \multicolumn{1}{l}{mAP} \\
		\hline
		w/o partition     & 83.93 & 79.04 & 53.46 & 52.09 \\
		\hline
		5     & 86.22 & 82.53 & 58.22 & 59.33 \\
		6     & 87.45 & 84.85 & 60.27 & 59.40 \\
		7     & 85.45 & 81.79 & 57.33 & 56.27 \\
		\hline \hline
	\end{tabular}%
	\label{tab:mixup-number}%
	\vspace{-0.5cm}
\end{table}%

\textbf{Analysis of modality-adaptive mixup scheme.} Table~\ref{tab:mixup-policy} shows the influence of different mixup schemes. Fix refers to a fixed mixup ratio of 0.5 for interpolating RGB and IR images, Beta refers to a mixup ratio randomly sampled from Beta prior distribution. MAM refers to our modality-adaptive mixup scheme. Fix and Beta belong to data-independent mixup schemes, while MAM is data-dependent. We can observe that the data-independent mixup schemes Fix and Beta obtain performance degradation over the data-dependent modality-adaptive mixup scheme. The comparison demonstrates that the proposed MAM can learn the dynamical modality discrepancy between different pair of RGB and IR images, and adaptively adjust the mixup ratio to generate appropriate mixed modality images for promoting a more continuous modality-invariant latent space and extracting more effective RGB-IR representations.

The results in Table~\ref{tab:mixup-number} show the influence of different number $G$ of the partitioned local regions for the modality-adaptive mixup scheme on RegDB dataset. W/o partition refers to the employ of a global mixup ratio to interpolate the whole RGB and IR images. We can observe that local linear interpolation policy ($\boldsymbol{m} \in \mathbb{R}^G$) obtains obvious performance improvement over global linear interpolation policy ($\boldsymbol{m} \in \mathbb{R}^1$), which indicates that adaptively mixing different local regions of RGB and IR images could reduce intra-modality discrepancy, and produce high-quality mixed modality images with identity consistency and fewer confusing information (\textit{e.g.}, occlusion, blurring and ghosting) for promoting cross-modality feature learning. Moreover, Table~\ref{tab:mixup-number} shows that when adopting 6 partitioned local regions, the proposed MID with $\boldsymbol{m} \in \mathbb{R}^6$ obtains the best  re-identification performance.

\begin{table}[!t]
	\small
	\centering
	\caption{Influence of different number of the decomposed residual blocks $n_d$.}
	\begin{tabular}{c|cccc}
		\hline \hline
		\multicolumn{1}{c|}{\multirow{2}[0]{*}{Mixup Scheme}} & \multicolumn{2}{c}{RegDB} & \multicolumn{2}{c}{SYSU-MM01} \\ 
		& \multicolumn{1}{l}{r1} & \multicolumn{1}{l}{mAP} & \multicolumn{1}{l}{r1} & \multicolumn{1}{l}{mAP} \\
		\hline
		0     & 83.43 & 79.13 & 56.22 & 57.12 \\
		1     & 85.94 & 82.69 & 56.71 & 58.45 \\
		2     & 86.71 & 83.54 & 57.52 & 58.49 \\
		3     & 87.45 & 84.85 & 60.27 & 59.40 \\
		4     & 84.38 & 82.27 & 56.75 & 57.25 \\
		5     & 82.58 & 80.99 & 54.51 & 56.58 \\
		\hline \hline
	\end{tabular}%
	\label{tab:number-decomposed}%
\end{table}%

\textbf{Analysis of modality-adaptive convolution decomposition.} Table~\ref{tab:number-decomposed} shows the influence of the number of residual blocks equipped with modality-adaptive convolution decomposition. From Table~\ref{tab:number-decomposed}, we can observe that when $n_d$ increases from 0 to 3, the proposed method obtains 4.02\% and 4.05\% improvements in rank-1 accuracy on RegDB and SYSU-MM01 datasets, respectively. When $n_d$ increases from 3 to 5, the performance decreases by 4.87\% and 5.76\% rank-1 accuracy on RegDB and SYSU-MM01 datasets. MID obtains the best performance when $n_d=3$, indicting that suitable decomposed residual blocks can capture more invariant visual semantics and improve the capacity of the learned modality-shared representations.



\begin{figure}[!t]
	\centering
	\includegraphics[width=0.47\textwidth]{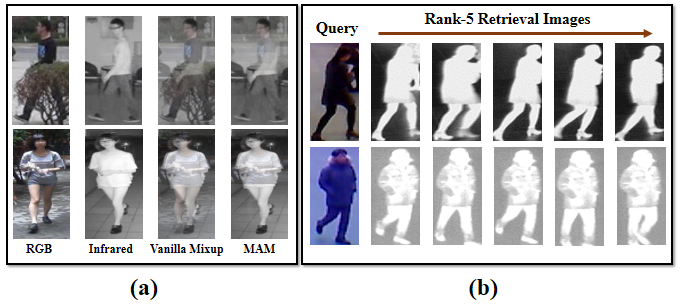}
	\centering
	\caption{(a) Visualization of the generated mixed modality images by the vanilla mixup scheme and the modality-adaptive mixup scheme; (b) Visualization of some retrieval results on RegDB dataset.}
	\label{fig:visual}
	\vspace{-0.5cm}
\end{figure}

\textbf{Visualization of results.} To further verify the benefit of the modality adaptive mixup scheme, we visualize some mixed modality images in Figure~\ref{fig:visual}(a). Compared with the vanilla mixup scheme \cite{zhang2017mixup}, the modality adaptive mixup scheme is able to generate high-quality mixed modality images without the interference of occlusion, blurring and ghosting, \textit{etc}. Moreover, we visualize some retrieval results in Figure~\ref{fig:visual}(b), which demonstrate the effectiveness of the proposed MID for identifying the same pedestrians and distinguishing different pedestrians.



\section{Conclusions}
In this work, we propose a novel modality-adaptive mixup and invariant decomposition (MID) approach for RGB-infrared person re-identification to mitigate the inherent modality discrepancy between RGB and IR images at the pixel-level and feature-level. MID firstly introduces modality-adaptive mixup scheme to generate appropriate mixed modality images by the actor-critic agent for reducing modality gap at the pixel-level and facilitating a more continuous modality-invariant latent space. MID then designs modality-adaptive convolution decomposition to simultaneously counter modality discrepancy and enforce cross-domain shared semantics at the feature-level, for learning effective modality-shared representation. Experimental results on two cross-modality person re-identification datasets have demonstrated the superiority of the proposed method.

\section{Acknowledgments}
This work was supported by the National Key R\&D Program of China under Grant 2020AAA0105702, National Natural Science Foundation of China (NSFC) under Grant U19B2038 and 62106245, the University Synergy Innovation Program of Anhui Province under Grants GXXT-2019-025, and the Fundamental Research Funds for the Central Universities under Grant WK2100000021.
\bibliography{aaai22.bib}

\end{document}